# A Cumulative Multi-Niching Genetic Algorithm for Multimodal Function Optimization

Matthew Hall
Department of Mechanical Engineering
University of Victoria
Victoria, Canada

***Abstract*—This paper presents a cumulative multi-niching genetic algorithm (CMN GA), designed to expedite optimization problems that have computationally-expensive multimodal objective functions. By never discarding individuals from the population, the CMN GA makes use of the information from every objective function evaluation as it explores the design space. A fitness-related population density control over the design space reduces unnecessary objective function evaluations. The algorithm's novel arrangement of genetic operations provides fast and robust convergence to multiple local optima. Benchmark tests alongside three other multi-niching algorithms show that the CMN GA has greater convergence ability and provides an order-of-magnitude reduction in the number of objective function evaluations required to achieve a given level of convergence.**

## I. INTRODUCTION

Genetic algorithms provide a powerful conceptual framework for creating customized optimization tools able to navigate complex discontinuous design spaces that could confound other optimization techniques. In this paper, I present a new genetic algorithm that uniquely combines two key capabilities: high efficiency in the number of objective function evaluations needed to achieve convergence, and robustness in optimizing over multi-modal objective functions. I created the algorithm with these capabilities to meet the needs of a very specific optimization problem: the design of floating platforms for offshore wind turbines. However, the algorithm's features make it potentially valuable for any application that features a computationally-expensive objective function and multiple local optima in a discontinuous design space.

Many design optimization problems have computationally-expensive objective functions. While genetic algorithms (GAs) may be ideal optimizers in many ways, a conventional GA's disposal of previously-evaluated individuals from past generations constitutes an unnecessary loss of information. Rather than being discarded, these individuals could instead be retained and used to both inform the algorithm about good and bad regions of the design space and prevent the redundant evaluation of nearly-identical individuals. This could accelerate the optimization process by significantly reducing the number of objective function evaluations required for convergence to an optimal solution.

Examples in the literature of GA approaches that store previously-evaluated individuals in memory to reduce unnecessary or redundant objective function evaluations are sparse. Xiong and Schneider [1] developed what they refer to as a Cumulative GA, which retains all individuals with a high fitness value to use along with the current generation in reproduction. This approach is useful in retaining information about the best regions of the design space, but it does nothing to avoid redundant objective function evaluations. A GA developed by Gantovnik et al. [2], however, does. Their GA stores information about all previous individuals and uses it to construct a Shepard's method response surface approximation of surrounding fitness values, which can be used instead of evaluating the objective function for nearby individuals.

Retaining past individuals to both provide information about the design space and avoid redundant objective function evaluations was my first goal in developing a new GA. My second goal was for the algorithm to be able to identify and converge around multiple local optima in an equitable way.

Identifying multiple local optima is necessary for many practical optimization problems that have multimodal objective functions. Even though an objective function may have only one global optimum, another local optimum may in fact be the preferred choice once additional factors are considered – factors that may be too complex, qualitative, or subjective to be included in the objective function. In the optimization of floating offshore wind turbine platforms, for example, a number of distinct locally-optimal designs exist, ranging from wide barges to deep slender spar-buoys. Though a spar-buoy may have the greatest stability (a common objective function choice), a barge design may be the better choice once ease of installation is considered.

Furthermore, global optimizations often use significant modelling approximations in the objective function for the sake of speed in exploring large design spaces. It is possible for such approximations to skew the design space such that the wrong local optimum is the global optimum in the approximated objective function. In those cases, local gradient-based optimizations with higher-fidelity models in the objective function are advisable as a second optimization stage to verify the locations of the local optima and determine which one of them is in fact the global optimum.

A conventional GA will only converge stably to one local optimum but a number of approaches have been developed for enabling convergence to multiple local optima, a capability referred to as "multi-niching". The Sharing approach, proposed by Holland [3] and expanded by Goldberg and Richardson [4], reduces the fitness of each individual based on the number of neighbouring individuals. The fitness reduction is determined by a sharing function, which includes a threshold distance that determines what level of similarity constitutes a neighbouring individual. A weakness of this approach is that choosing a good sharing function requires a-priori knowledge of the objective function characteristics. As well, the approach has difficulty in forming stable sub-populations, though improvements have been made in this area [5].

An alternative is the Crowding approach of De Jong [6], which features a replacement step that determines which individuals will make up the next generation: for each offspring, a random subset of the existing population is selected and from it the individual most similar to the offspring is replaced by it. Mahfoud's improvement, called Deterministic Crowding [7], removes the selection pressure in reproduction by using random rather than fitness-proportionate selection, and modifies the replacement step such that each crossover offspring competes against the more similar of its parents to decide which of the two enters the next generation.

The Multi-Niche Crowding approach of Cedeño [8] differs from the previous crowding approaches by implementing the crowding concept in the selection stage. For each crossover pair, one parent is selected randomly or sequentially and the other parent is selected as the most similar individual out of a group of randomly selected individuals.

This promotes mating between nearby individuals, providing stability for multi-niching. The replacement operation is described as "worst among most similar"; a number of groups are created randomly from the population, the individual from each group most similar to the offspring in question is selected, and the least fit of these "most similar" individuals is replaced by the offspring.

Though the Multi-Niche Crowding approach is quite effective at finding multiple local optima, it and the other approaches described above still provide preferential treatment to optima with greater fitness values. Lee, Cho, and Jung provide another approach, called Restricted Competition Selection [9], that outperforms the previously-mentioned techniques in finding and retaining even weak local optima. In their otherwise-conventional approach, each pair of individuals that are within a "niche radius" of each other are compared and the less fit individual's fitness is set to zero. This in effect leaves only the locally-optimal individuals to reproduce. A set of the fittest of these individuals is retained in the next generation as elites.

Some more recent GAs add the use of directional information to provide greater control of the design space exploration. Hu et al. go so far as to numerically calculate the gradient of the objective function at each individual in order to use a steepest descent method to choose offspring [10].

This approach is powerful, but its large number of function evaluations makes it impractical for computationally-expensive objective functions. Liang and Leung [11] use a more restrained approach in which two potential offspring are created along a line connecting two existing individuals and the four resulting fitness values are compared in order to predict the locations of nearby peaks. By using this information to inform specially-constructed crossover and mutation operators, this algorithm uses significantly fewer function evaluations than other comparable GAs [11].

An approach shown to use even fewer function evaluations is an evolutionary algorithm (EA) by Cuevas and Gonźalez that mimics collective animal behaviour [12]. This algorithm models the way animals are attracted to or repelled from dominant individuals, and retains in memory a set of the fittest individuals. Competition between individuals that are within a threshold distance is also included. Notwithstanding the lack of a crossover function, this algorithm is quite similar in operation to many of the abovementioned GAs and is therefore easily compared with them. It is noteworthy because of its demonstrated efficiency in terms of number of objective function evaluations.

None of the abovementioned multi-niching algorithms retains information about all the previously-evaluated individuals; a GA that combines this sort of memory with multi-niching is a novel creation. In developing such an algorithm, which I refer to as the Cumulative Multi-Niching (CMN) GA, I drew ideas and inspiration from many of the abovementioned approaches. In some cases, I replicated specific techniques, but in different stages of the GA process. The combination of genetic operations to make up a functioning GA is entirely unique.

## II. Algorithm Description

The most distinctive feature of the CMN GA is that it is cumulative. Each successive generation adds to the overall population. With the goal of minimizing function evaluations, evaluated individuals are never discarded; even unfit individuals are valuable in telling the algorithm where not to go. The key to making the cumulative approach work is the use of an adaptive proximity constraint that prevents offspring that are overly similar to existing individuals from being added to the population. By using a distance threshold that is inversely proportional to the fitness of nearby individuals, the CMN GA encourages convergence around promising regions of the design space and allows only a sparse population density in less-fit regions of the design space.

This fundamental difference from other GAs enables a number of unique features in the genetic operations of the algorithm that together combine (as summarized in Fig. 2) to make the cumulative multi-niching approach work. The selection and crossover operations are designed to support stable sub-populations around local optima and drive the algorithm's convergence. The mutation operation is designed to encourage diversity and exploration of the design space. The "addition" operation, which takes the place of the replacement operation of a conventional GA, is designed to make use of the accumulated population of individuals in order to avoid redundant or unnecessary fitness function evaluation and guide the GA to produce offspring in the most promising regions of the design space. The fitness scaling operation makes the GA treat local optima equally despite potential differences in fitness. The details of these operations are as follows.

### *A. Selection and Crossover*

The selection and pairing process for crossover combines fitness-proportionate selection with a crowding-inspired pairing scheme that is biased toward nearby individuals. Whereas Cedeño's Multi-Niche Crowding approach selects the first parent randomly and selects its mate as the nearest of a randomly-selected group, the CMN GA combines factors of both fitness and proximity in its selection operation.

The first parent, P1, of each pair is selected from the population using standard fitness-proportionate selection (FPS) – with the probability of selection proportional to fitness. Then, for each P1, a crowd of $N_{crowd}$ candidate mates is selected using what could be called proximity-proportionate selection (PPS) - with the probability of selection determined by a proximity function describing how close each potential candidate mate, P2, is to P1 in the design space. The most basic proximity function is the inverse of the Euclidean distance:

$$P_{P1,P2} = \frac{1}{\sqrt{\sum_{i=1}^{n}\left(X_i^{P1} - X_i^{P2}\right)^2}} \quad (1)$$

where X is an individual's decision variable vector, with length n. The fittest of the crowd of candidate mates is then selected to pair with P1. This process is repeated for each individual selected to be a P1 parent for crossover.

By having an individual mate with the fittest of a crowd of individuals that are mostly neighbours, mating between members of the same niche is encouraged, though the probability-based selection of the crowd allows occasional mating with distant individuals, providing the important possibility of crossover between niches. This approach contributes to the CMN GA's multi-niching stability and is the basis for crossover-driven convergence of the population to local optima.

In the crossover operation, an offspring's decision variable values are selected at uniform random from the hypercube bounded by the decision variable values of the two parents.

### *B. Mutation*

The mutation operation occurs in parallel with the crossover operation. Mutation selection is done at random, and the mutation of the decision variables of each individual is based on a normal distribution about the original values with a tuneable standard deviation. This gives the algorithm the capability to widely explore the design space. Though individual fitness is not explicitly used in the mutation operation, the addition operation that follows makes it more likely that mutations will happen in fitter regions of the design space.

### *C. Addition*

The cumulative nature of the CMN GA precludes the use of a replacement operation. Instead, an addition operation adds offspring to the ever-expanding population. A proximity constraint ensures that the algorithm converges toward fitter individuals and away from less fit individuals. This filtering, which takes place before the offspring's fitnesses are evaluated, is crucial to the success of the cumulative population approach. By rejecting offspring that are overly similar to existing members of the population, redundant objective function evaluations are avoided.

The proximity constraint's distance threshold, $R_{min}$, is inversely related to the fitness of the nearest existing individual, $F_{nearest}$, as determined by a distance threshold function. A simple example is:

$$R_{min} = 0.1\,(1.01 - F_{nearest}) \quad (2)$$

This function results in a distance threshold of 0.001 around the most fit individual and 0.101 around the least fit individual, where distance is normalized by the bounds of the design space and fitness is scaled to the range [0 1].

This approach for the addition function allows new offspring to be quite close to existing fit individuals but enforces a larger minimum distance around less fit individuals. As such, the population density is kept high in good regions and low in poor regions of the design space, as determined by the accumulated objective function evaluations over the course of the GA run. A population density map is essentially prescribed over the design space as the algorithm progresses. If the design space was known a priori, the use of a grid-type exploration of the design space could be more efficient, but without that knowledge, this more adaptive approach is more practical.

To adjust for the changing objectives of the algorithm as the optimization progresses – initially to explore the design space and later to narrow in on local optima - the distance threshold function can be made to change with the number of individuals or generation number, G. This can help prevent premature convergence, ensuring all local optima are identified. The distance threshold function that I used to generate the results in this paper is:

$$R_{min} = 0.08\left[1.001 - F_{nearest}\left(1 - 0.5(0.9)^G\right)\right] \quad (3)$$

### *D. Fitness Scaling*

The algorithm described thus far could potentially converge to only the fittest local optimum and not adequately explore other local optima. The final component, developed to resolve this problem and provide equitable treatment of all significant local optima, is a proximity-weighted fitness scaling operation. In most GAs, a scaling function is applied to the population's fitness values to scale them to within normalized bounds and also sometimes to adjust the fitness distribution. A basic approach is to linearly scale the fitness values, F, to the range [0, 1] so that the least fit individual gets a scaled fitness of F'=0 and the fittest individual gets a scaled fitness of F'=1:

$$F'_i = \frac{F_i - \min(F)}{\max(F) - \min(F)} \quad (4)$$

A scaling function can also be used to adjust the distribution of fitness across the range of fitness values in order to, for example, provide more or less emphasis on moderately-fit individuals. This scaling can be adaptive to the characteristics of the population. For the results presented here, I used a second, exponential scaling function to adjust the scaled fitness values so that the median value is 0.5:

$$F''_i = (F'_i)^{\left[\frac{\ln(0.5)}{\ln(\text{median}(F'))}\right]} \quad (5)$$

Proximity-weighted fitness scaling, a key component of the CMN GA, adds an additional scaling operation. This operation relies on the detection of locally-optimal individuals in the population. The criterion I used, for simplicity, is that an individual is considered to represent a local optimum if it is fitter than all of its nearest $N_{min}$ neighbours. In the proximity-weighted fitness scaling operation, scaling functions (4) and (5) are applied multiple times to the population, each time normalizing the results to the fitness of a different local optimum. So if m local optima have been identified, each individual in the population will have m scaled fitness values. These scaled fitness values F'' are then combined for each individual i according to the individual's proximity to each respective local optimum j to obtain the population's final scaled fitness values:

$$F'''_i = \frac{\sum_{j=1}^{m} P_{i,j} F''_{i,j}}{\sum_{j=1}^{m} P_{i,j}} \quad (6)$$

Proximity, $P_{i,j}$, can be calculated as in (1). This process gives each local optimum an equal scaled fitness value, as is illustrated for a one-dimensional objective function in Fig. 1.

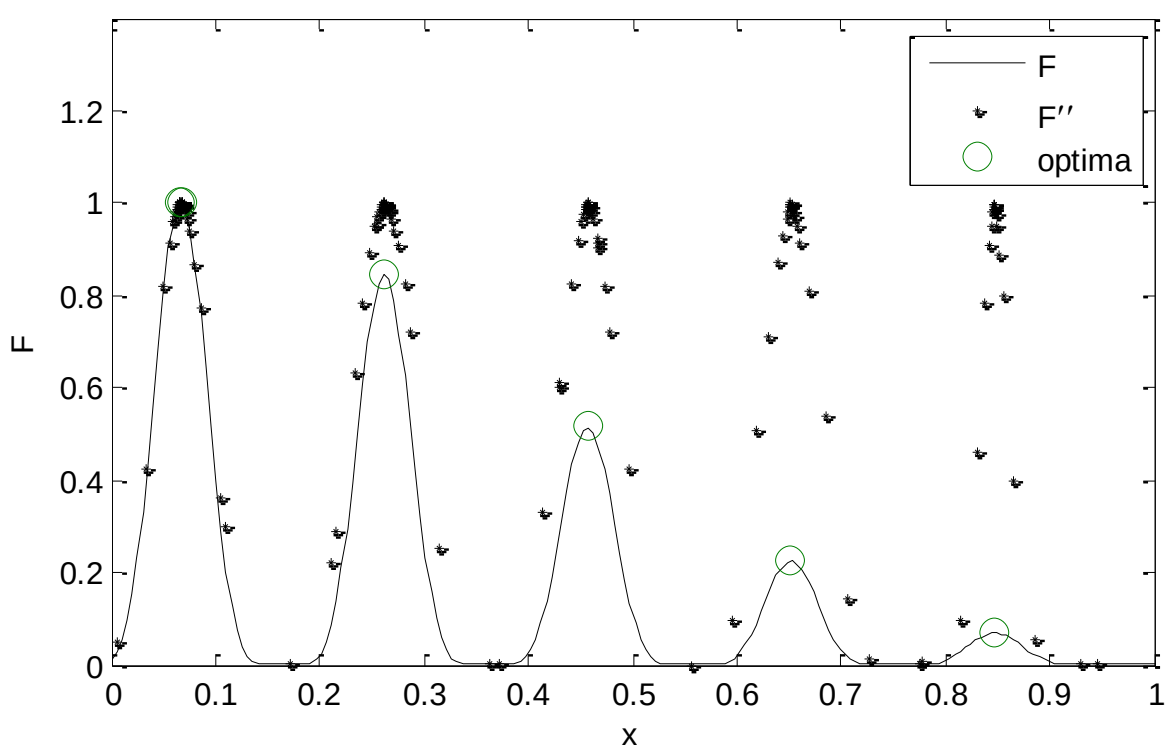

Figure 1. Proximity-weighted fitness scaling.

### *E. CMN GA Summary*

Fig. 2 describes the overall structure of the CMN GA, outlining how the algorithm's operations are ordered and how the addition operation filters out uninformative offspring. The next section demonstrates the algorithm's effectiveness at multi-niche convergence with a minimal number of objective function evaluations.

## III. PERFORMANCE RESULTS

To benchmark the CMN GA's performance, I tested it alongside three other multi-niching algorithms on four generic multimodal objective functions. These four multimodal functions have been used by many of the original developers of multi-niching GAs [8].

```
Step 0: (Initialization)
  Randomly generate Npop individuals
  Evaluate the individuals' fitnesses F

Step 1: (Fitness Scaling)
  Calculate distances between individuals
  Identify locally-optimal individuals
  For each individual i:
    For each locally-optimal individual j:
        Calculate scaled fitness F″i,j
    Calculate proximity-weighted fitness F‴i

Step 2: (Crossover)
  Select a P1 from the population using FPS
  Select a crowd of size Ncrowd using PPS
  Select the fittest in the crowd to be P2
  Cross P1 and P2 to produce an offspring
  If offspring satisfies distance threshold:
    Add to population and calculate fitness F
  Repeat Ntry times or until Ncrossover offspring
  have been added to the population

Step 3: (Mutation)
  Randomly select a mutation individual
  Mutate individual to produce an offspring
  If offspring satisfies distance threshold:
    Add to population and calculate fitness F
  Repeat Ntry times or until Nmutate offspring
  have been added to the population

Step 4: (New Generation)
  Repeat from Step 1 until stopping criterion
  is met
```

Figure 2. CMN GA outline.

The first, F1, is a one-dimensional function featuring five equal peaks, shown in Fig. 3.

$$F_1(x) = \sin^6(5.1\,\pi x + 0.5) \quad (7)$$

The second, F2, modifies F1 to have peaks of different heights, shown in Fig. 4.

$$F_2(x) = \exp\left(-\frac{4(ln2)(x-0.0667)^2}{0.64}\right) F_1(x) \quad (8)$$

The third, F3, is a two-dimensional Shekel Foxholes function with 25 peaks of unequal height, spaced 16 units apart in a grid, as shown in Fig. 5.

$$F_3(x,y) = 0.002 + \sum_{i=1}^{25} \frac{1}{i+(x-A_i)^6+(y-B_i)^6} \quad (9)$$

The fourth, F4, is an irregular function with five peaks of different heights and widths, as listed in Table 1 and shown in Fig. 6.

$$F_4(x,y) = \sum_{i=1}^{5} \frac{H_i}{1+W_i[(x-A_i)^2+(y-B_i)^2]} \quad (10)$$

In F3 (9) and F4 (10), $A_i$ and $B_i$ are the x and y coordinates of each peak. In F4 (10), $H_i$ and $W_i$ are the height and width parameters for each peak. These four functions test the algorithms' multi-niching capabilities in different ways.

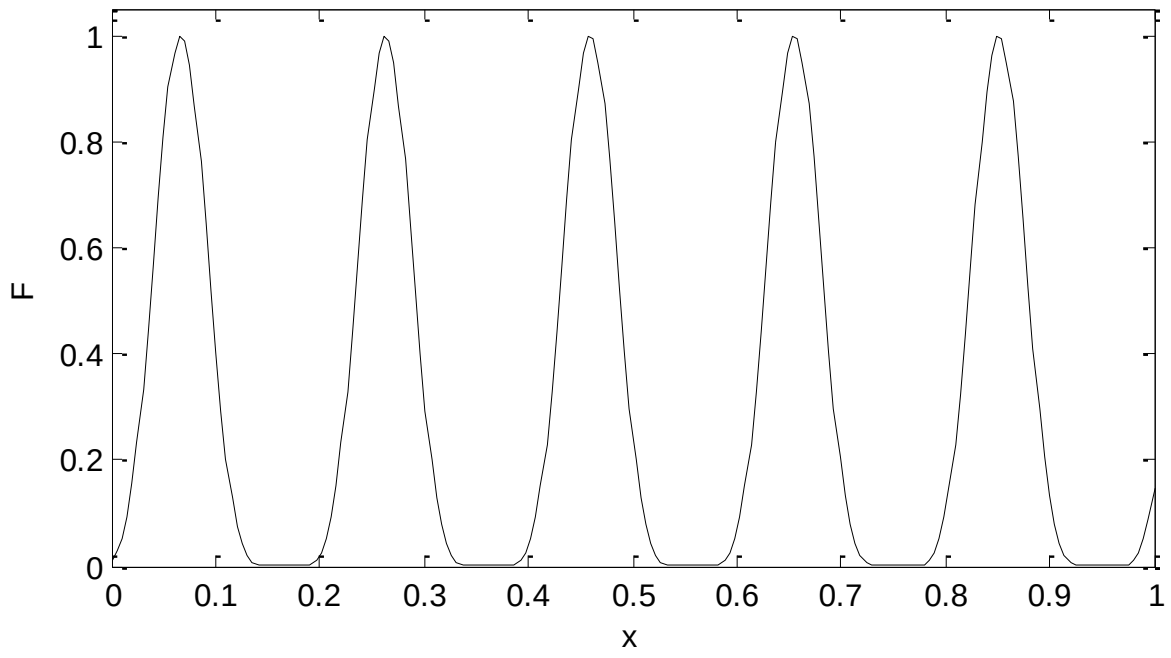

Figure 3. F1 objective function.

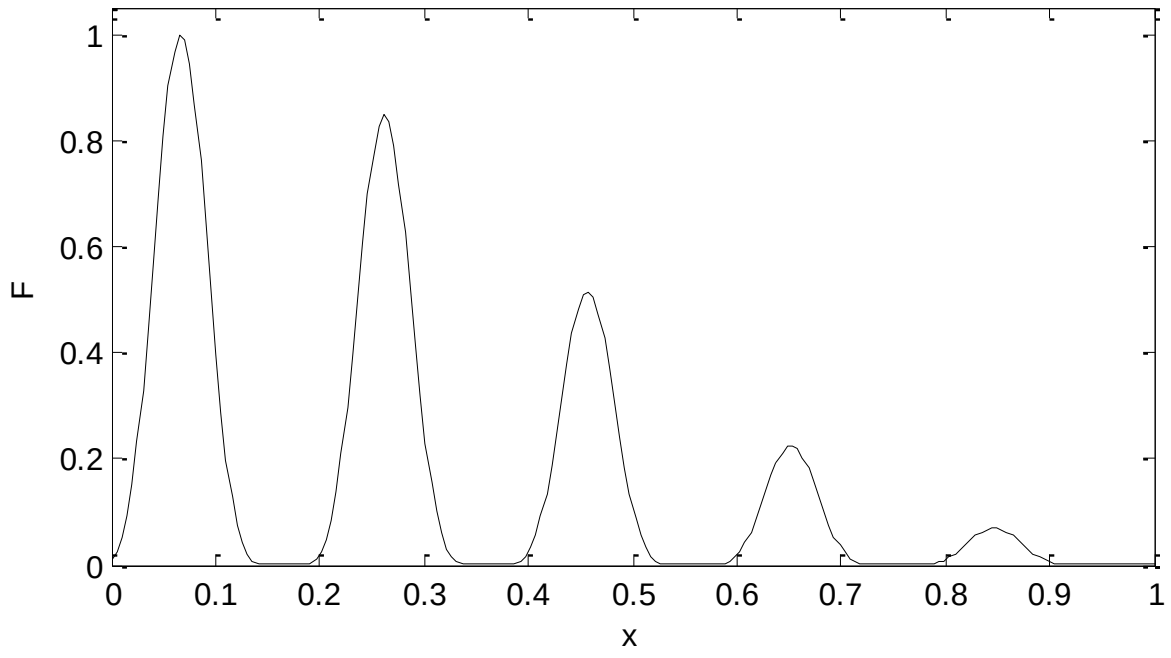

Figure 4. F2 objective function.

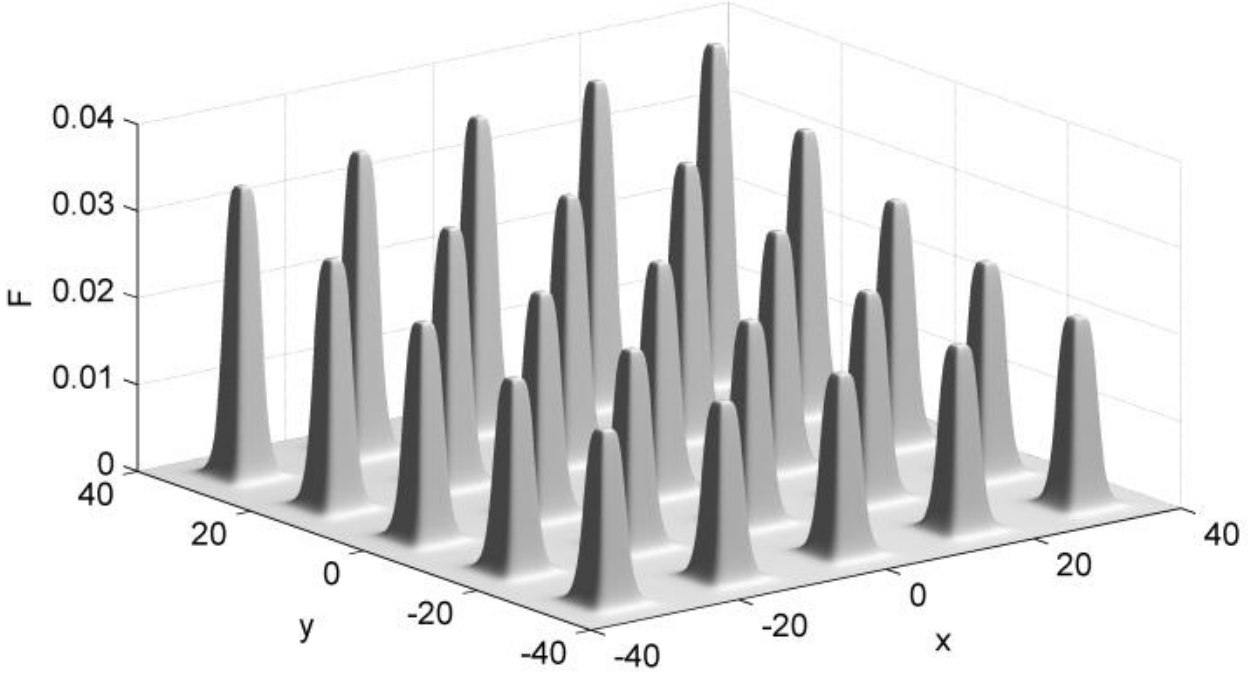

Figure 5. F3 objective function.

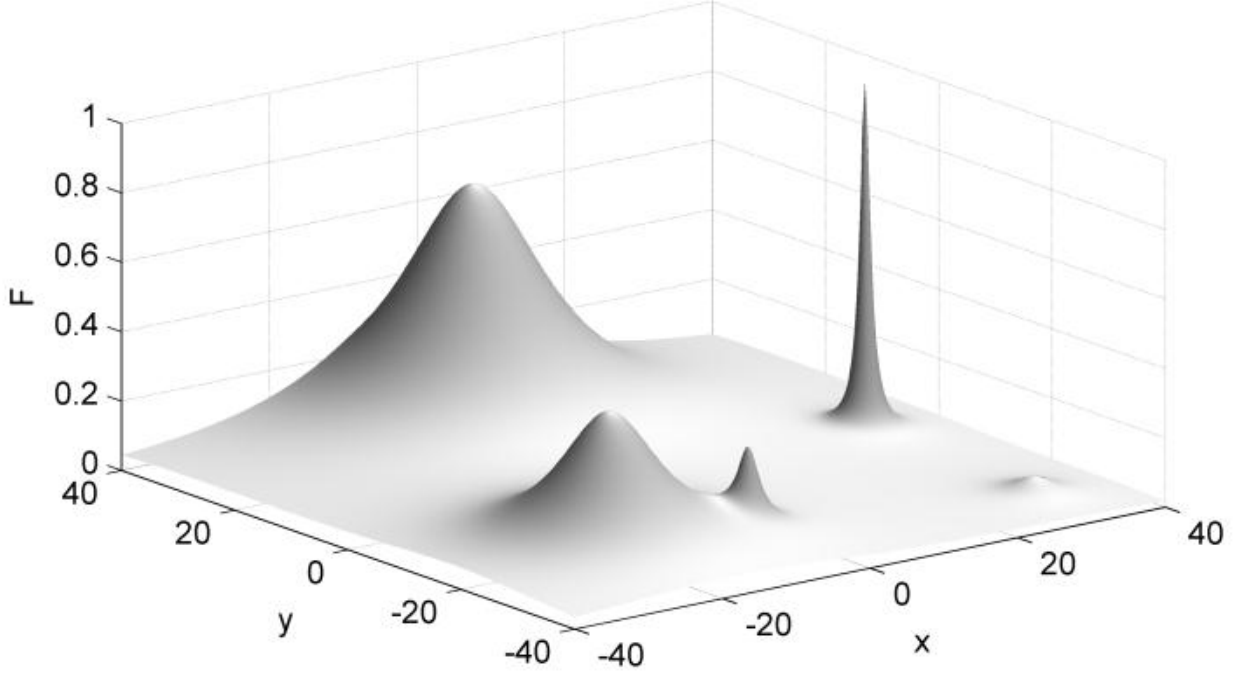

Figure 6. F4 objective function.

TABLE I. F4 OBJECTIVE FUNCTION PEAKS

| I | $A_i$ | $B_i$ | $H_i$ | $W_i$ |
|---|---|---|---|---|
| 1 | -20 | -20 | 0.4 | 0.02 |
| 2 | -5 | -25 | 0.2 | 0.5 |
| 3 | 0 | 30 | 0.7 | 0.01 |
| 4 | 30 | 0 | 1.0 | 2.0 |
| 5 | 30 | -30 | 0.05 | 0.1 |

The two other multi-niching GA approaches I compare the CMN GA against are Multi-Niche Crowding (MNC) [8] and Restricted Competition Selection (RCS) [9]. I chose these two because they are very well-performing examples of two different approaches to GA multi-niching. I implemented these techniques into a GA framework that is otherwise the same as the CMN GA in terms of how it performs the crossover and mutation operations.

Crossover offspring decision variable values are chosen at uniform random from the intervals between the decision variables of the two parents. Mutation offspring decision variable are chosen at random using normal distributions about the unmutated values with standard deviations of 40% of the design space dimensions.

For further comparison, I also implemented the Collective Animal Behaviour (CAB) evolutionary algorithm [12]. It is a good comparator because it has many common features with multi-niching GAs, but has been shown to give better performance than many of them, particularly in terms of objective function evaluation requirements.

The values of the key tunable parameters used in each algorithm are given in Tables 2 to 5. $N_{pop}$ describes the population size, or the initial population size in the case of the CMN GA. For the RCS GA, $N_{elites}$ is the number of individuals that are preserved in the next generation. I tuned the parameter values heuristically for best performance on the objective functions. For the MNC, RCS, and CAB algorithms, I began by using the values from [8], [9], and [12], respectively, but found that modification of some parameters gave better results. The meanings of the variables in Table 4 can be found in [12].

To account for the randomness inherent in the operation of a genetic or evolutionary algorithm, I ran each algorithm ten times on each objective function to obtain a reliable characterization of performance. The metric I use to measure the convergence of the algorithms to the local optima is the sum of the distances from each local optimum $X^*_j$ to the nearest individual.

By indicating how close the algorithm is to identifying all of the true local optima, this aggregated metric represents what is of greatest interest in multimodal optimization applications. The assumption is that in real applications it will be trivial to determine which evaluated individuals represent local optima without a-priori knowledge of the objective function.

TABLE II. PARAMETERS FOR THE MNC GA TECHNIQUE

| Function | F1 & F2 | F3 & F4 |
|---|---|---|
| $N_{pop}$ | 50 | 200 |
| $N_{crossover}$ | 45 | 180 |
| $N_{mutation}$ | 5 | 20 |
| $C_S$ | 15 | 75 |
| $C_F$ | 3 | 4 |
| $S$ | 15 | 75 |

TABLE III. PARAMETERS FOR THE RCS GA TECHNIQUE

| Function | F1 & F2 | F3 & F4 |
|---|---|---|
| $N_{pop}$ | 10 | 80 |
| $N_{elites}$ | 5 | 30 |
| $N_{crossover}$ | 8 | 50 |
| $N_{mutation}$ | 2 | 30 |
| $R_{niche}$ | 0.1 | 12 |

TABLE IV. PARAMETERS FOR THE CAB EA TECHNIQUE

| Function | F1 & F2 | F3 & F4 |
|---|---|---|
| $N_{pop}$ | 20 | 200 |
| $B$ | 10 | 100 |
| $H$ | 0.6 | 0.6 |
| $P$ | 0.8 | 0.8 |
| $v$ | 0.01 | 0.001 |
| $\rho$ | 0.1 | 4 |

TABLE V. PARAMETERS FOR THE CMN GA TECHNIQUE

| Function | F1 & F2 | F3 & F4 |
|---|---|---|
| $N_{pop\ (initial)}$ | 10 | 100 |
| $N_{crossover}$ | 3 | 20 |
| $N_{mutation}$ | 2 | 12 |
| $N_{min}$ | 3 | 6 |
| $N_{crowd}$ | 10 | 20 |
| $N_{try}$ | 100 | 100 |

Figures 7 to 10 show plots of the convergence metric versus the number of objective function evaluations for each optimization run. Using these axes gives an indication of algorithm performance in terms of my two objectives for the CMN GA, convergence to multiple local optima and minimal objective function evaluations. Figures 7, 8, 9, and 10 compare the performance of each algorithm for objective functions F1, F2, F3, and F4, respectively.

In the results for objective function F4, the MNC and CAB algorithms consistently failed to identify the shallowest peak. Accordingly, I excluded this peak from the convergence metric calculations for these algorithms in the data of Fig. 10 in order to provide a more reasonable view of these algorithms' performance. The CMN GA also missed this peak in one of the runs, as can by the one anomalous curve in Fig. 10, wherein the convergence metric stagnates at a value of 2. As is the case with other multi-niching algorithms, missing subtle local optima is a weakness of the CMN GA, but it can be mitigated

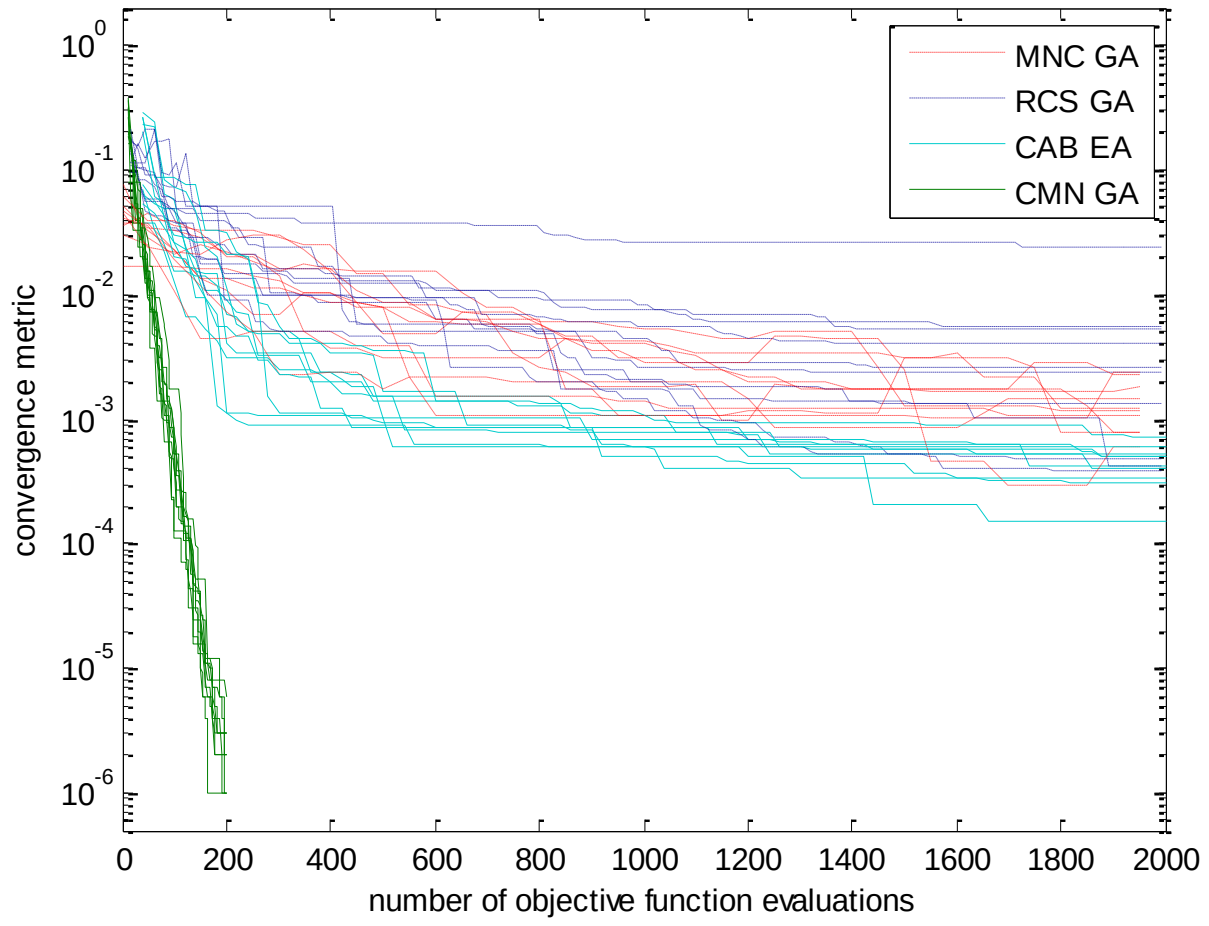

Figure 7. GA performance for F1 objective function runs.

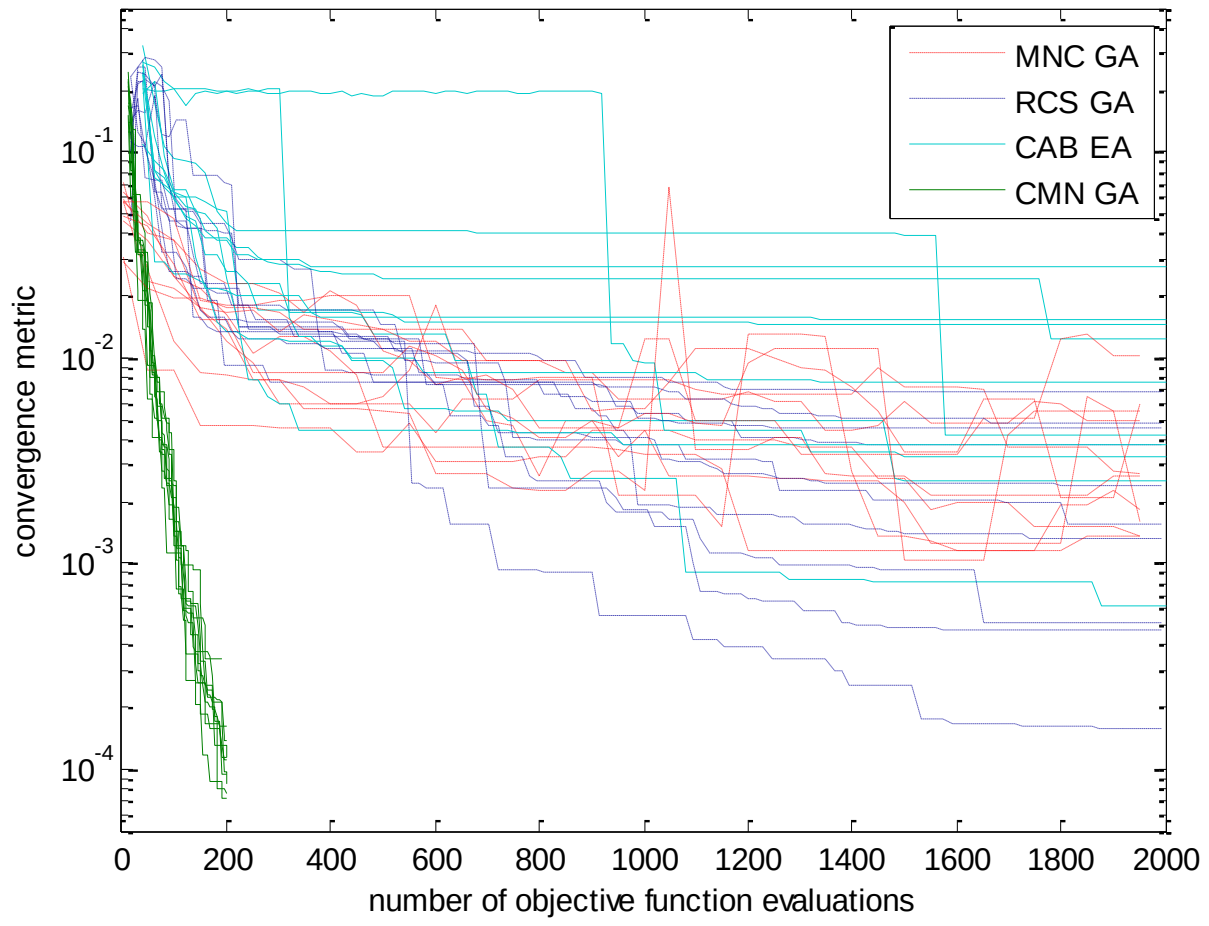

Figure 8. GA performance for F2 objective function runs.

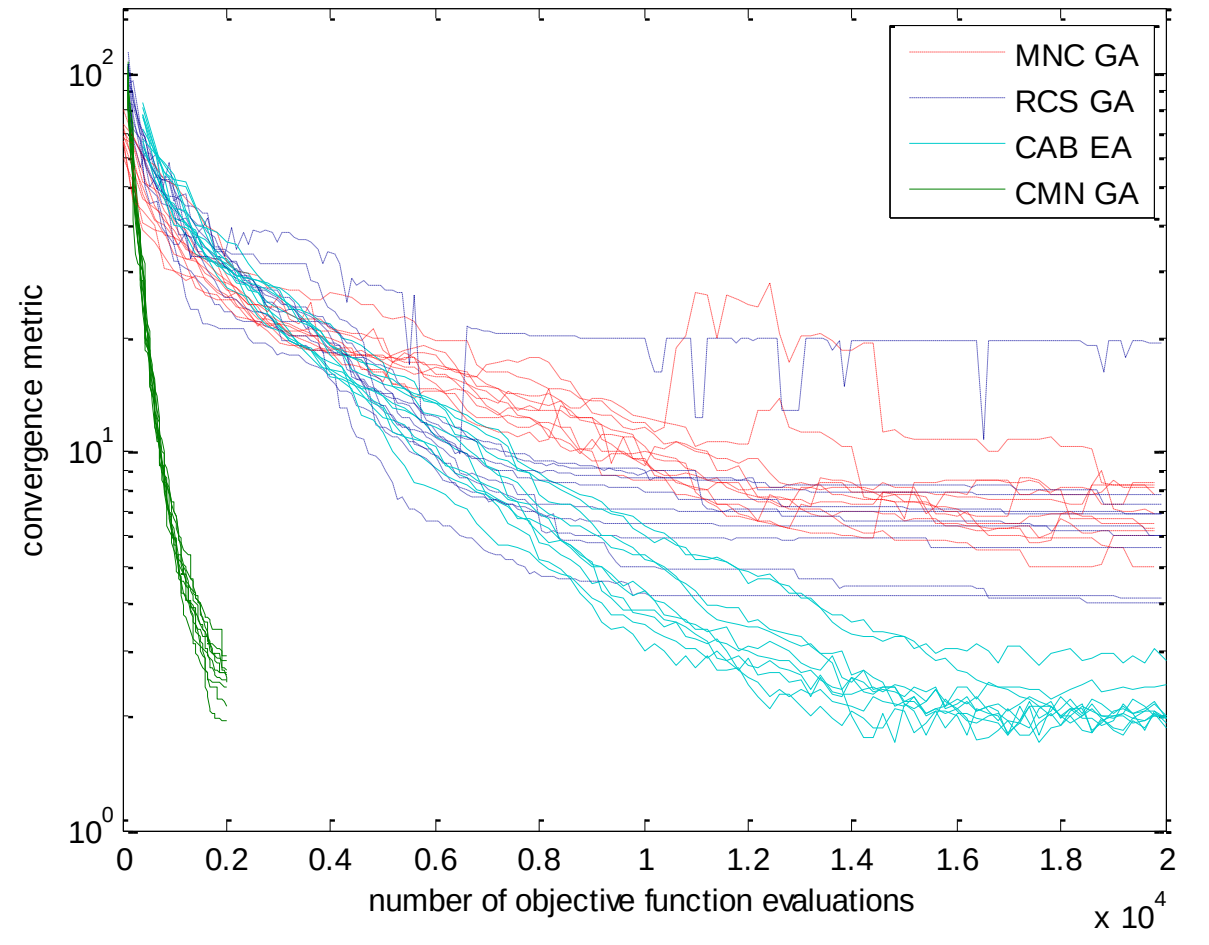

Figure 9. GA performance for F3 objective function runs.

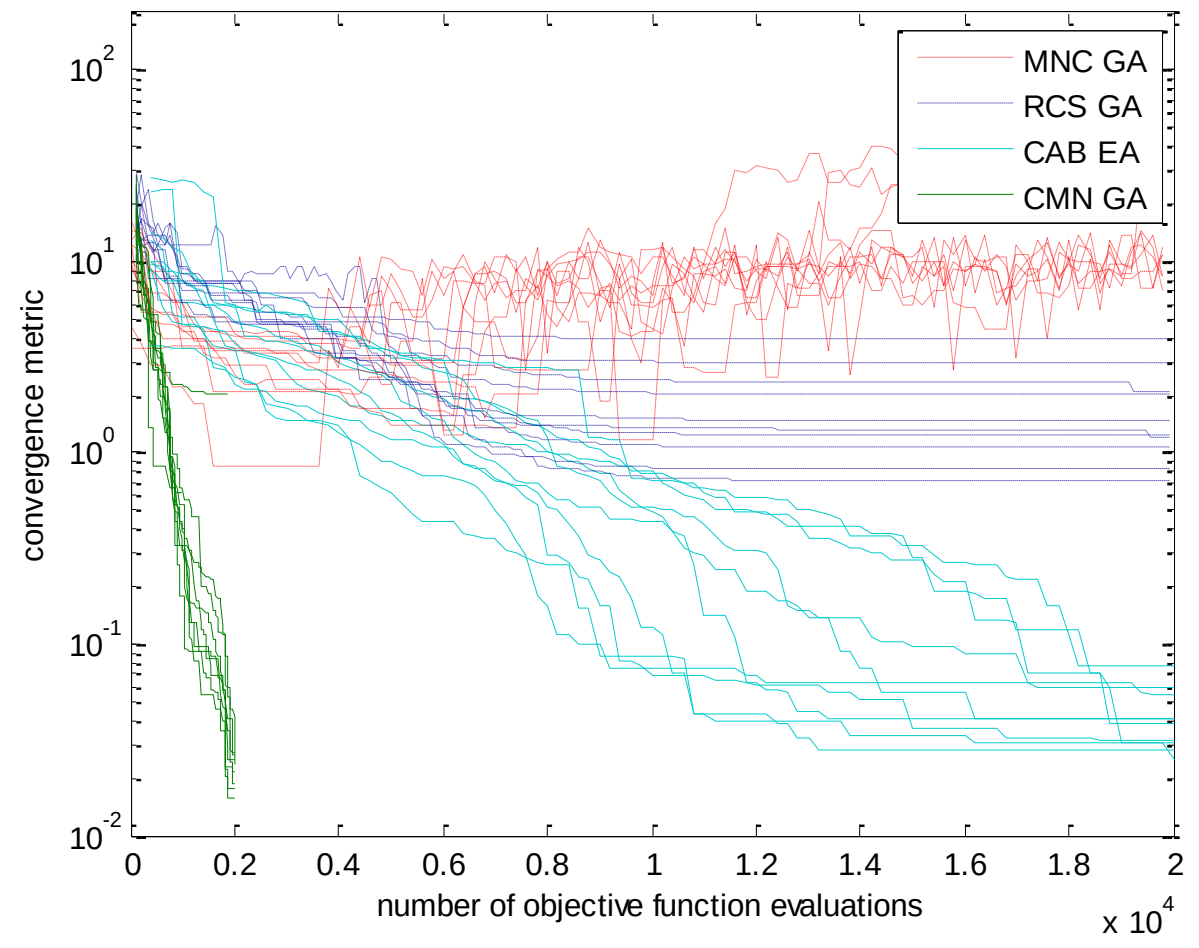

Figure 10. GA performance for F4 objective function runs.

by careful choice of algorithm parameters and verifying results through multiple optimization runs.

Fig. 11 is a snapshot of a population generated by the CMN GA on the F4 objective function. The distribution of the 1000 individuals in the figure illustrates how the algorithm clearly identifies the five local optima and produces a high population density around them regardless of how shallow or sharp they may be. Fig 12 shows how, with the same input parameters, the CMN GA is just as effective with the 25 local optima of the F3 objective function.

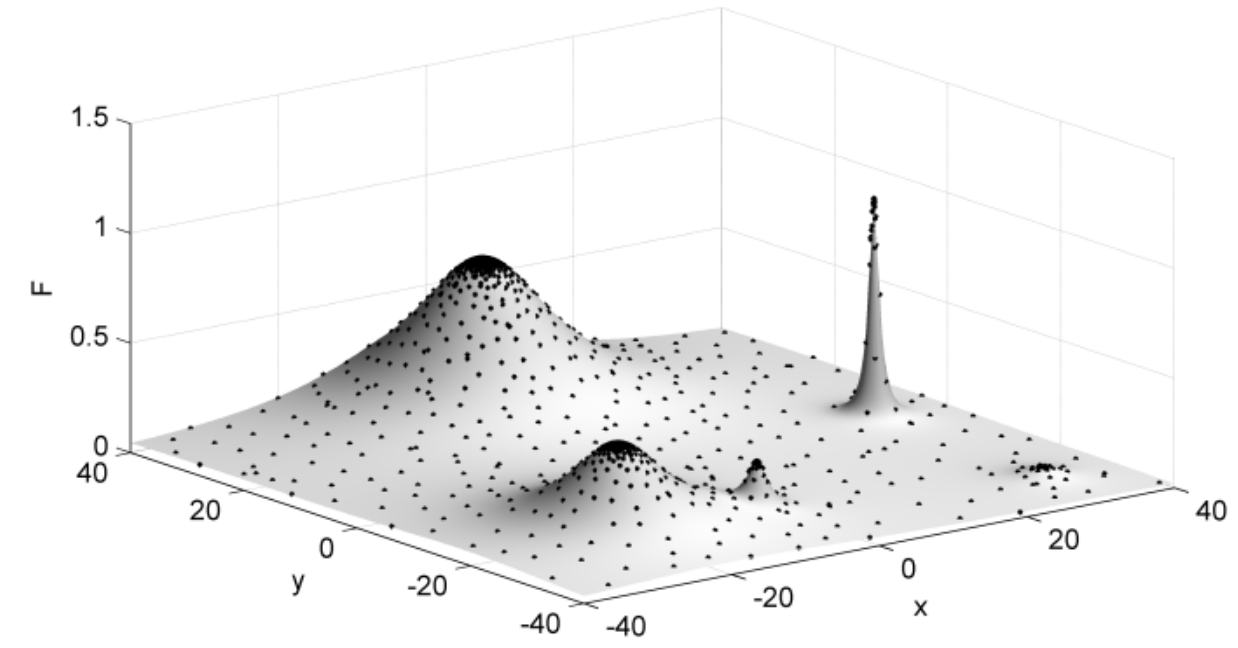

Figure 11. CMN GA exploration of F4 objective function.

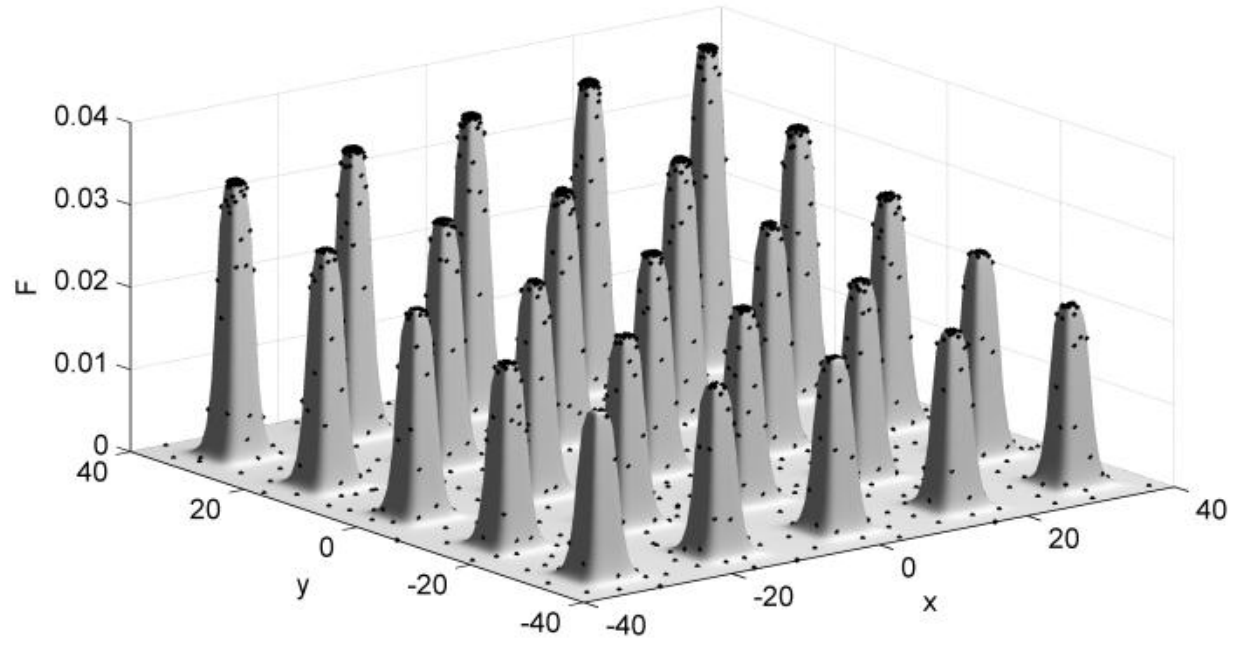

Figure 12. CMN GA exploration of F3 objective function.

Though more rigorous tuning of parameters could result in slight performance improvements in any of the four algorithms I compared, the order-of-magnitude faster convergence of the CMN GA gives strong evidence of its superior performance in terms of multimodal convergence versus number of objective function evaluations.

It should be noted that this measure of performance, reflective of the design goals of the CMN GA, is only indicative of performance on optimization problems where evaluating the objective function dominates the computational effort. The algorithm operations of the CMN GA are themselves much slower than those of the other algorithms, so the CMN GA could be inferior in terms of computation time on problems with easily-computed objective functions. As well, with its ever-growing population, the CMN GA's memory requirements are greater than those of the other algorithms. In a sense, my choice of measure of performance puts the MNC, RCS, and CAB algorithms at a disadvantage because, unlike the CMN GA, these algorithms were not designed specifically for computationally-intensive objective functions. That said, convergence versus number of function evaluations is the most relevant measure of performance for optimizing over computationally-expensive multimodal objective functions, and the algorithms I chose for comparison represent three of the best existing options out of the selection of applicable GA/EA approaches available in the literature.

## IV. Conclusion

In the interest of efficiently finding local optima in computationally-expensive objective functions, I created a genetic algorithm that converges robustly to multiple local optima with a comparatively small number of objective function evaluations. It does so using a novel arrangement of genetic operations in which new individuals are continuously added to the population; I therefore call it a Cumulative Multi-Niching Genetic Algorithm. The tests presented in this paper demonstrate that the CMN GA meets its goals – convergence to multiple local optima with minimal objective function evaluations – strikingly better than alternative genetic or evolutionary algorithms available in the literature. It therefore represents a useful new capability for optimization problems that have computationally-expensive multimodal objective functions. The proximity constraint approach used to control the accumulation of individuals in the population may also be applicable to other metaheuristic algorithms.